# *Amalgam*: A Framework for Obfuscated Neural Network Training on the Cloud

Sifat Ut Taki
University of Notre Dame
Indiana, USA
staki@nd.edu

Spyridon Mastorakis
University of Notre Dame
Indiana, USA
mastorakis@nd.edu

## ABSTRACT

Training a proprietary Neural Network (NN) model with a proprietary dataset on the cloud comes at the risk of exposing the model architecture and the dataset to the cloud service provider. To tackle this problem, in this paper, we present an NN obfuscation framework, called *Amalgam*, to train NN models in a privacy-preserving manner in existing cloud-based environments. *Amalgam* achieves that by augmenting NN models and the datasets to be used for training with well-calibrated noise to "hide" both the original model architectures and training datasets from the cloud. After training, *Amalgam* extracts the original models from the augmented models and returns them to users. Our evaluation results with different computer vision and natural language processing models and datasets demonstrate that *Amalgam*: (i) introduces modest overheads into the training process without impacting its correctness, and (ii) does not affect the model's accuracy. The prototype implementation is available at: https://github.com/SifatTaj/amalgam

## CCS CONCEPTS

## KEYWORDS

## 1 INTRODUCTION

The rise of machine learning (ML) applications has driven a surge in the need for high-performance computing resources [1, 43]. Although hardware like GPUs and TPUs can supercharge training speed for Neural Networks (NN), it's tough for users to get their hands on them directly [37]. Many cloud providers have allowed access to such accelerators (e.g., Google Colab, Kaggle Notebooks, JetBrains Datalore, etc.). These public services are becoming more popular as ML-based applications are becoming mainstream. Most of these services provide access to a Python-based environment with either queue-based free access or subscription-based premium access to machine learning accelerators. Owning a private cloud instance can be significantly costlier than public cloud services[1].

While training an NN model on the cloud is a popular alternative to acquiring expensive hardware and training it locally, it poses a new challenge. When a user designs a proprietary NN model architecture and trains it with a proprietary dataset on the cloud, the cloud service provider has access to the dataset and the NN model architecture information [19]. This raises privacy concerns for users and organizations when training a proprietary NN model or using a proprietary dataset to train a model on the cloud. Furthermore, there are no straightforward mechanisms to prevent that, since the cloud environment needs to have access to the NN model architecture, parameters, and training dataset to properly train a model [3].

Preserving the model's privacy when training on the cloud is a major challenge to achieve [14, 49, 56]. A model training framework running on the cloud needs to have access to the model architectures, parameters, and training dataset to train a model successfully. As such, this problem cannot be solved by simply "hiding" the model parameters from cloud providers. Solutions like homomorphic encryption (HE) and multi-party computation can be slow and complex to manage [13, 16, 50]. Moreover, the existing privacy-preserving frameworks are not designed for public clouds.

[1]Google Colab Pro costs $9.90 per month whereas a similarly configured VM can cost up to $453.5 per month at the time of writing this paper.

In this paper, to tackle this challenge, we propose *Amalgam*: a framework for obfuscated NN model training on the cloud. *The main novelty of Amalgam is that it "augments" the NN model architecture and dataset in ways that hide the original model architecture and dataset information from the cloud while training, without impacting the correctness of the training process and the performance of the trained model (*e.g., *convergence of loss and accuracy during training).* Furthermore, *Amalgam* is designed from the ground up to work with any Python-based ML cloud services (e.g., Google Colab, GC Vertex AI, AWS Sagemaker, etc.), which no other privacy-preserving NN training framework aims to achieve (to the best of our knowledge). Although it incurs a modest overhead in terms of training time and memory, we find this trade-off worthwhile when preserving the privacy of both the model architecture and the training dataset. The trained model can be used for inference using the original dataset. *Amalgam* also allows users to fine-tune a pre-trained model. Moreover, obfuscation is proven to provide privacy for deployed neural networks [57].

*Amalgam aims to reduce the complexity of training NN models in an obfuscated manner on the cloud.* As such, *Amalgam* does not require a specialized cloud environment and can run on any Python-based environment, since it uses the *PyTorch* library as its backbone. In other words, any NN model defined in *PyTorch* can be obfuscated through *Amalgam*. *Amalgam* can also be deployed locally on user devices, such as laptops and desktop computers. Our contributions to this paper are the following:

- We present the design (Sections 3 and 4), implementation (Section 5), as well as security and privacy analysis (Section 6) of *Amalgam*. We describe how augmentation is performed to achieve the obfuscation of both the NN model architecture and the training dataset without affecting the training outcomes of the original model.
- We present evaluation results using widely-used NN models and datasets for computer vision and natural language processing (Section 5). Our results also include a performance comparison with other frameworks for privacy-preserving model training. Finally, we performed several adversarial attacks against *Amalgam* to demonstrate its resilience against these attacks (Section 6).

## 2 RELATED WORK

**Secure Multi-party Computation (MPC):** Secure MPC is one of the most popular approaches to train NN models in a secure environment. SecureNN [45] proposed by Wagh et al. leverages three-party secure computation for neural network training by secretly sharing the stochastic gradient descent procedure when training a model. Agrawal et al. implemented a two-party secure neural network training and prediction technique called QUOTIENT [2]. XONN [38] proposed an xor-based multiplication approach to reduce the computation overhead in matrix multiplications. Finally, CrypTFlow [25] by Kumar et al. proposed a three-party semi-honest MPC-based secure framework for TensorFlow inference. MPC-based approaches remain impractical since they require significant computation power and are complex to deploy and maintain.

**Homomorphic Encryption (HE):** HE allows training NNs on encrypted data, keeping the data and the model confidential. As such, the information remains private in an untrusted environment. Nandakumar et al. achieved NN training on encrypted data and model information using Fully Homomorphic Encryption (FHE) [36]. Lee et al. realized a scheme based on FHE for NN model training that allows arithmetic operations on encrypted real and complex numbers [28]. This approach, however, is impractical as it requires approximately four hours for a single inference on a dual Intel Xeon Platinum 8280 CPU (112 cores) with 512 GB of memory. DOReN was proposed by Meftah et al. to address the problem with space efficiency through a batched neuron that can evaluate multiple quantized neurons on encrypted data without approximations [33]. However, the authors did not evaluate a full NN. Overall, solutions based on HE remain impractical for NN model training and inference in their current state.

**Federated Learning (FL):** FL offers a different approach to training neural networks while preserving privacy compared to HE by training an NN model collaboratively on multiple devices without ever sharing the raw data itself. [32]. FL trains a global model by training smaller local models distributed among different devices or organizations–eliminating the need to access the datasets used by the local models and only needing to exchange trained parameters [35]. The goal of *Amalgam*, however, is different, since it aims to address privacy concerns when a model is trained on the cloud based on a dataset provided by the user.

**Differential Privacy (DP):** DP in neural network training focuses on protecting individual data points within the training dataset by adding well-calibrated noise [12]. Various DP techniques, such as metric DP [7], local DP [11], shuffled DP [6], and hybrid DP [4], have been proposed. When DP is applied to machine learning, a model $\mathcal{M}$ is trained on a dataset $\mathcal{D}$ producing a model in space $R$ using the $\mathcal{M} : \mathcal{D} \rightarrow \mathcal{R}$ mechanism [31]. However, DP has an impact on model training, since the accuracy of the smaller classes drops significantly [5].

**Table 1: Overview of different privacy-preserving frameworks.**

| Technique | Usability | Overhead | Accuracy loss | GPU acceleration | Compatibility |
|---|---|---|---|---|---|
| SMPC | Complex | High | No | Yes | All models |
| HE | Simple | Very High | Yes | No | Limited models |
| FL | Complex | Medium | Yes | Yes | All models |
| DP | Simple | High | Yes | Yes | Limited datasets |
| TEE | Complex | High | No | No | Limited models |
| *Amalgam* | Simple | Low | No | Yes | All models |

**Trusted Execution Environment (TEE):** TEE-based solutions (e.g., TensorScone [26], Plinius [51]) typically incur substantial overhead during overhead or even a drop in accuracy for large models/datasets. For example, TensorScone is limited to using CPU for training while Plinius struggles with large models/datasets due to the limited enclave page cache size resulting in extensive page swaps. Moreover, TEE-based solutions require support from the underlying hardware architecture. On the other hand, *Amalgam* can run on any Python-based environment.

**How is *Amalgam* different than prior work:** The goal of *Amalgam* is to obfuscate both an NN architecture and a training dataset when training on a cloud platform. With *Amalgam*, users do not have to deploy and maintain their own cloud instances. *Amalgam* works with any model or dataset (without any modifications from the users) on any available Python-based cloud training services–requiring no special environment and using GPUs for training without any impact on accuracy. Finally, *Amalgam* does not impact the training procedure of the original network so that the original network can be used for inference with the original dataset after training. Table 1 presents the properties and comparison of different privacy-preserving frameworks.

## 3 THREAT MODEL AND DESIGN OVERVIEW

**Threat model:** In *Amalgam*, we consider a cloud provider as an attacker. As the model is defined and trained with a dataset in the environment of a cloud provider, it has access to the model architecture and the dataset [19]. Side-channel attacks can be also performed by monitoring the resource utilization while training a model [48, 59] on remote cloud computing resources. We also considered various types of attacks discussed in Section 6.3. As of today, there are no straightforward ways to train a model in a privacy-preserving manner on the cloud as the cloud environment needs to have access to the model parameters and the dataset features. As such, proprietary NN models and datasets are vulnerable to attacks when trained remotely on the cloud.

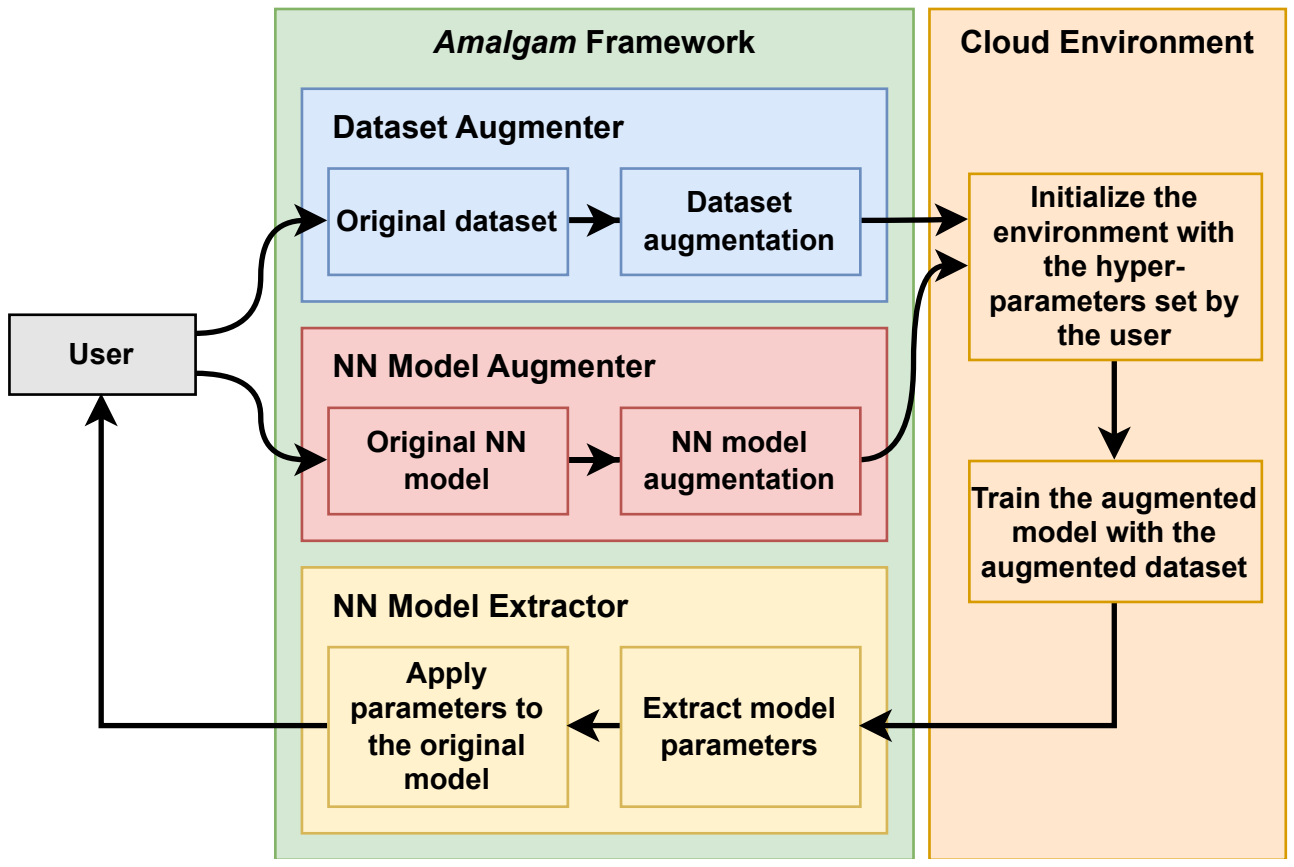

**Figure 1: Workflow of the *Amalgam* operation.**

**Design overview:** *Amalgam* can run on user devices (*e.g.,* laptop or desktop computers). The framework consists of the following modules: (1) Dataset Augmenter, (2) NN Model Augmenter, and (3) NN Model Extractor. Figure 1 presents the workflow of *Amalgam*. First, a user defines the NN architecture (model) and provides a dataset to train. The model augmenter then augments the model to obfuscate its architecture and characteristics through the addition of synthetic noise in the form of additional layers, parameters, and connections. Subsequently, the dataset augmenter augments the provided dataset through the addition of noise (*e.g.,* synthetic pixels for image datasets). The augmented model and dataset are sent to the cloud environment for training. After training, *Amalgam* receives the augmented model from the cloud, extracts the original model from the augmented model using the NN model extractor module, and returns the original model to the user.

## 4 SYSTEM ARCHITECTURE

### 4.1 Dataset Augmenter

This module is responsible for the obfuscation of a dataset. Users provide a dataset for model training, along with an augmentation amount (as a percentage) and the type of noise to be added to the dataset. The augmentation amount determines the quantity of noise to obfuscate the data. The noise added to the original dataset does not alter the original information, and it conceals the characteristics without impacting the training validity (*e.g.,* accuracy, loss). Dataset augmentation can be done with any type of image or text data. The dataset is first processed into *tensors* of numeric values with one or more dimensions. The dataset augmenter inserts synthetic numeric values (depending on the noise type) into each sample of data randomly when pre-processing a dataset. We describe the dataset augmentation process with examples from computer vision (image datasets) and Natural Language Processing (NLP) (text datasets) below.

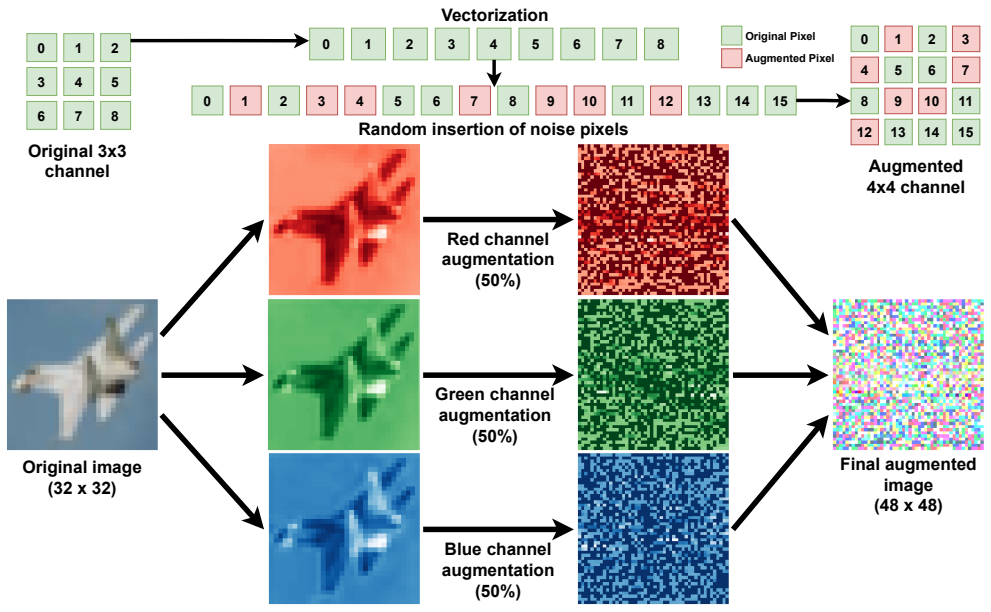

**Figure 2: Image augmentation process of *Amalgam* along with an example with three channels from the CIFAR10 dataset.**

Users can choose a noise type from three categories: random noise (default option), Gaussian/Laplace noise, and user-provided noise. The random noise is chosen *uniformly at random* from the range of a maximum possible value and a minimum possible value. The Gaussian/Laplace noise is randomly chosen from either a Gaussian or a Laplace distribution based on $\sigma$. Furthermore, users can provide their own noise data to augment the dataset. For example, a user may use pixels from actual meaningful images to augment a dataset. In that case, the augmented images will contain meaningful pixels from the provided noise. Note that the noise augmentation is not trivial, thus they cannot simply be denoised from augmented datasets (further discussed in Section 6.3).

**Image Dataset Augmentation:** Image datasets are obfuscated through the augmentation of additional pixels as noise. When training a Convolutional Neural Network (CNN) model, an RGB image is processed into a 3D *tensor*, which is passed to the first convolution layer of the model. The dataset augmenter accepts a dataset $\{d_1, d_2, d_3, \ldots, d_n\} \in \mathcal{D}$ and then it inserts synthetic numeric values $\{x_a, y_a\}$ as noise into random indices of the processed *tensors* of $\{d'_1, d'_2, d'_3, \ldots, d'_n\} \in \mathcal{D}'$, where $\mathcal{D}'$ is the new augmented dataset (obfuscated). To augment an item $d$ into $d'$, the *tensors* are first separated into their corresponding channels so that they become 2D *tensors* for each channel. Subsequently, each *tensor* is vectorized and augmented with a random array of synthetic pixels $\{x_a, y_a\}$. Finally, all three 2D *tensors* are combined to produce the final augmented image *tensor* $d'$. The final shape of the augmented *tensor* $d'$ depends on the augmentation amount $A_d$, which the user selects. As a result, the final dimension of an image with a dimension of $X \times Y$ will be $(X + (X \times A_d)) \times (Y + (Y \times A_d))$. For example, if an original image is 10x10 in dimension then it would have 100 indices. If the target augmented dimension is 11x11 then Amalgam would insert 21 synthetic pixels into random indices, making the total size of indices 121. Figure 2 presents a simplified example of image augmentation (dimension of 3×3), along with an example image from the *CIFAR10* dataset [24], which has been augmented with 50% augmentation. The original image size is 32×32, which becomes 48×48 after obfuscation.

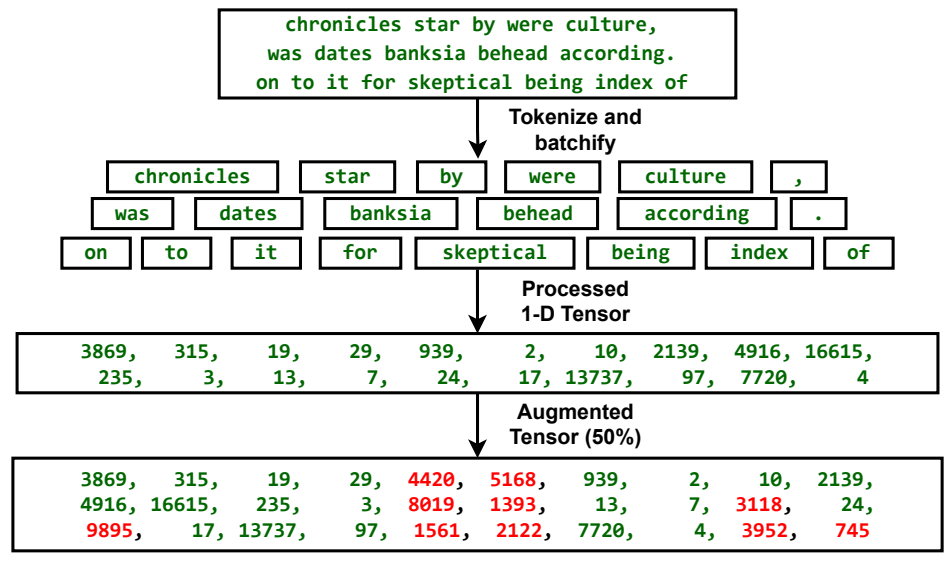

**Figure 3: A text sample augmentation example from the WikiText2 dataset.**

**Text Dataset Augmentation:** Text datasets are obfuscated via the augmentation of synthetic tokens. When training an NLP model with a dataset $\mathcal{T}$ containing text samples $\{t_1, t_2, t_3, \ldots, t_n\} \in \mathcal{T}$, the text data is first tokenized and vectorized into a 1D *tensor*. The *tensor* is then divided into multiple batches with a pre-defined batch size. The dataset augmenter inserts synthetic numeric values (depending on the type of noise) at random indices into the pre-processed 1D *tensor* to generate a new dataset of $\{t'_1, t'_2, t'_3, \ldots, t'_n\} \in \mathcal{T}'$, where $\mathcal{T}'$ is the new augmented text dataset. The user selects the augmentation amount $A_d$, and the final length of each batch becomes $X + (X \times A_d)$ from $X$. Figure 3 shows an example from the *WikiText2* dataset [34], which has been augmented with 50% noise. In the future, we will make *Amalgam* compatible with other types of datasets as well (e.g., audio data).

## 4.2 NN Model Augmenter

Model obfuscation is performed by the NN model augmenter component. For this component, a user provides a model, an augmentation amount, and the type of noise for augmentation. Amalgam generates $n_s$ sub-networks with synthetic parameters that also contain connections to other sub-networks, which creates branches of networks. This applies to both CV and NLP models with all types of layers. The sub-network containing the original layers does not receive input from other augmented layers; however, some outputs from the original layers go to other augmented layers. Since the original layers only send output to other augmented layers and do not receive input from other augmented layers, the accuracy of the original model remains the same. it makes the process of tracing the original layers significantly difficult. By default, it generates a random number of sub-networks; however, users may specify it.

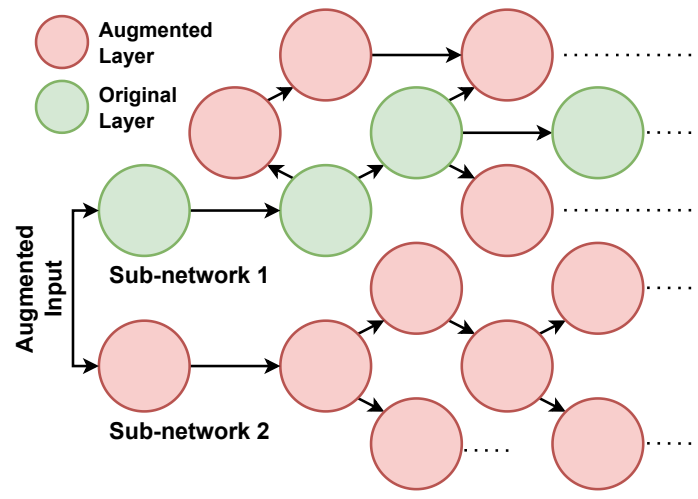

**Figure 4: A simplified model augmentation example with 2 sub-networks.**

Each sub-network $\{m_1, m_2, \ldots, m_s\}$ receives the entire augmented input but processes random information from an augmented dataset $\mathcal{D}'$. The original sub-network processes the original inputs. As such, the training accuracy of the original model is not affected. For example, if an augmented image has 9 pixels, and each sub-network processes a subset of 4 pixels. The sub-network containing the original layers will process the original 4 pixels, whereas the other sub-networks will process random subsets of 4 pixels. The randomized subsets can be overlapping and repeating. One subset may go to multiple sub-networks. Even the original subset may go to multiple sub-networks based on randomization. This makes the identification of the original sub-network even with model explanation techniques infeasible (further discussed in Section 6.3). Figure 4 presents an example of model augmentation with 2 sub-networks.

**CNN Augmentation:** After augmenting a CNN model $\mathcal{M}$ with synthetic parameters and generating an augmented model $\mathcal{M}'$ with multiple sub-networks, the NN model augmenter component selects the first input layers $\{l_1', l_1'', \ldots, l_1^s\}$ of each sub-network and replace them with $\{c_1', c_1'', \ldots, c_1^s\}$, where $c$ is a custom convolution layer designed for *Amalgam*. We designed a **custom convolution layer** for *Amalgam*, which is capable of skipping a set of inputs $(x_a, y_a)$ when running the convolution operation on each channel of the input image. It uses the following equation for each channel in the forward pass:

$$G_{(x,y)} = \sum_{\substack{\delta x=-k_i \\ \forall \delta x \notin x_a}}^{k_i} \sum_{\substack{\delta y=-k_j \\ \forall \delta y \notin y_a}}^{k_j} h_{out,in}(\delta x, \delta y).F(x+\delta x, y+\delta y), \quad (1)$$

where $k_i$ and $k_j$ are the pixel position in the kernel filter, $h$ is the output of the next layer, and $F$ is the filter. An augmented layer $l_a'$ can have inputs from another augmented layer $l_b'$ or from an original layer $l_b$. Subsequently, multiple sub-networks are created within the final augmented model.

**Algorithm 1** *Amalgam* training algorithm

**Input:** Augmented data $\{d_1', d_2', \ldots, d_n'\} \in \mathcal{D}'$, augmented model $\mathcal{M}'$, loss function $\mathcal{L}$, learning rate $\eta$, batch T.
*Initialization* : Random $\theta_0^1, \theta_0^2, \ldots, \theta_0^s$

1: **for** $\forall t \in T$ **do**
2: $\quad g_t^s(d_i') \longleftarrow \nabla_{\theta_t^s} \mathcal{L}(\theta_t^s, d_i')$ for $\forall s \in \mathcal{M}'$
3: $\quad \theta_{t+1}^s \longleftarrow \theta_t^s - \eta_t g_t^s$

**Output:** Trained $\theta_T^1, \theta_T^2, \ldots, \theta_T^s$

**NLP Model Augmentation:** NLP models have different structures than CNN models. Most NLP models consist of encoders and decoders. Sequence-to-sequence networks using recurrent neural networks and transformer networks using multi-head self-attention mechanisms are two popular approaches for NLP applications [42, 46]. It is a common practice in NLP to use input embedding to convert plain text from a dataset to a set of *tensors* $\mathcal{T}$, which are used to train an NLP model [15]. Similar to CNN model augmentation, when an NLP model $M$ is also augmented with synthetic parameters to create $M'$, the NN model augmenter component identifies the first embedding layers $\{l_1', l_1'', \ldots, l_1^s\}$ of each sub-network of $\mathcal{M}'$. Subsequently, $\{l_1', l_1'', \ldots, l_1^s\}$ is replaced with **custom embedding layers** $\{e_1', e_1'', \ldots, e_1^s\}$ designed for *Amalgam*. The embedding layer works as a lookup table that stores embeddings of a fixed dictionary and size. The custom embedding layers for the NN model augmenter can be formulated as follows:

$$y = \sum_{\substack{i=1 \\ i \notin x_a}}^{L} W_2 \sigma(\langle x, W_1 \rangle), \quad (2)$$

where $W_1$ is the input matrix, $W_2$ is the output matrix, $L$ is the number of layers, and $x_a$ is a vector of indices that a custom embedding layer ignores when embedding words. Using Equation 2, the augmented NLP model can perform the embedding operation on selected inputs only. Similar to CNN models, the embedding layer of the original NLP model is replaced by a custom embedding layer that we discussed above. Subsequently, the original model is augmented with random layers, creating multiple sub-networks within the augmented model. Each sub-network has a custom embedding layer with synthetic $x_a'$, dimension values, and padding to hide the original embedding layer information.

The training process of *Amalgam* using an augmented model and an augmented dataset is presented in Algorithm 1.

## 4.3 NN Model Extractor

This component receives a trained augmented model $\mathcal{M}'$ from the cloud. To extract the original network, it first creates a new network $\mathcal{M}$ containing original layers $\{l_1, l_2, l_3, \ldots, l_n\}$ using the original model definition provided by the user.

Since *Amalgam* knows the original architecture, this component copies the original trained weights of $\{l_1, l_2, l_3, \ldots, l_n\} \in \mathcal{M}'$, and applies those weights to $\{l_1, l_2, l_3, \ldots, l_n\} \in \mathcal{M}$. $\mathcal{M}$ does not contain a custom convolution layer or a custom embedding layer. As such, $\mathcal{M}$ can work with the original dataset instead of the augmented dataset.

## 4.4 Transfer Learning/Fine-Tuning

Transfer learning or fine-tuning is important in machine learning, where a model trained on a certain task is repurposed for another related task [60]. Considering the volume of computing resources and time needed to train a model from scratch, transfer learning or fine-tuning a model has been a popular choice for users. This can reduce the required volume of computing resources and/or training time. *Amalgam* facilitates transfer learning as well. If a model contains pre-trained layers with pre-trained weights, the model augmentation does not impact the training on the pre-trained weights. After a user defines a model $\mathcal{M}$, the user can apply pre-trained weights to $\{l_1, l_2, l_3, \ldots, l_n\} \in \mathcal{M}$ (the model to be augmented) before passing it to *Amalgam*. *Amalgam* generates $\{l'_1, l'_2, l'_3, \ldots, l'_m\}$ and augments them, which does not impact $\{l_1, l_2, l_3, \ldots, l_n\}$. When the model finishes training on the cloud, the NN model extractor applies the trained weights from $\{l_1, l_2, l_3, \ldots, l_n\} \in \mathcal{M}'$ to the original layers $\{l_1, l_2, l_3, \ldots, l_n\} \in \mathcal{M}$ to complete the process.

## 4.5 Selecting Appropriate Augmentation

Users should select their desired augmentation amount(s) based on their obfuscation requirements, the desired degree of resilience against adversarial attacks, and the computing overhead they can afford. In other words, there is a trade-off between the degree of obfuscation and attack resilience that can be achieved and the incurred computing overhead. Larger augmentation amounts provide greater degrees of obfuscation; however, they come at the cost of higher computing overheads, which are proportional to the selected augmentation amounts (further discussed in Section 6). On the other hand, smaller amounts of augmentation come at the cost of lower computing overheads, but they may provide weaker obfuscation. Moreover, different amounts of augmentation provide different levels of resilience against different types of attacks (further discussed in Section 6.3).

# 5 EVALUATION

We evaluated *Amalgam* based on popular computer vision and NLP models and datasets. We evaluated the three components of *Amalgam* in terms of the incurred overhead as well as model training and validation metrics (loss and accuracy) for different augmentation amounts. The *Dataset Augmenter* component was evaluated with four popular image datasets and two popular text datasets. The *NN model augmenter* component was evaluated by training four CNN models and two NLP models with different augmentation amounts and with the augmented image and text datasets. We compared the training of the augmented models and augmented datasets with the original models and datasets to showcase that *Amalgam* does not impact the training process. Finally, we evaluated the *NN model extractor* component by testing its loss and accuracy after extraction with the original testsets of the corresponding original datasets. Model training was performed on a server equipped with an Intel Core i9-10900X CPU, 64 GB of RAM, and two NVIDIA RTX 3090 GPUs.

## 5.1 *Amalgam* Prototype Implementation

We implemented a prototype of *Amalgam* with the *PyTorch* library. The dataset augmenter is implemented using the *LibTorch* library in C++. This component accepts image and text datasets saved in a *PyTorch tensor* as input along with an augmentation amount. This component outputs an augmented dataset and saves it as a *PyTorch tensor*. The NN model augmenter accepts an NN model defined with *PyTorch* as input and augments it based on a given augmentation amount. The augmented model is then saved as a *TorchScript*. The augmented model and the augmented dataset can be uploaded to a cloud python-based environment for training. For our evaluation, we used a server running a *jupyter notebook* environment. The NN model extractor is also implemented with *PyTorch*. This component accepts a trained augmented *TorchScript* model from the cloud once training is complete, applies the original trained parameters to the original model definition from the NN model augmenter, and returns the trained original model as a *TorchScript* model to the user.

## 5.2 Dataset Augmenter Evaluation

**Setup:** For the evaluation of the dataset augmenter with image datasets, we selected MNIST [9], CIFAR10 [24], CIFAR100 [24], and Imagenette [21]. For the evaluation with text datasets, we selected WikiText2 [34] and AGNews [53]. The augmentation amounts used for each dataset are 25%, 50%, 75%, and 100%. We measured the time needed to augment each dataset with different augmentation amounts. We also measured the final size of the augmented dataset and the additional search space incurred by the augmentation–which can be a good indication of obfuscation. The search space represents the total size of the vectorized augmented sample for each original pixel or token. For example, if 2 pixels are augmented with 9 noise pixels, then the search space for the first pixel is 9 and the second pixel is 8, making the total search space 17.

**Table 2: Dataset augmentation results.**

| Dataset | Augmentation amount | Average time (s) | Resolution | Dataset Size | Search space |
|---|---|---|---|---|---|
| MNIST | 0% (Original) | – | 28×28 | 219.6 MB | – |
| | 25% | 28.6 | 35×35 | 343 MB | $1.00 \times e^{346}$ |
| | 50% | 41.6 | 42×42 | 493.9 MB | $3.62 \times e^{524}$ |
| | 75% | 56.8 | 49×49 | 672.3 MB | $8.57 \times e^{656}$ |
| | 100% | 74 | 56×56 | 878.1 MB | $1.22 \times e^{764}$ |
| CIFAR10 | 0% (Original) | – | 32×32 | 737.6 MB | – |
| | 25% | 86.8 | 40×40 | 1.2 GB | $6.86 \times e^{452}$ |
| | 50% | 125 | 48×48 | 1.7 GB | $1.21 \times e^{686}$ |
| | 75% | 170.4 | 56×56 | 2.3 GB | $9.86 \times e^{858}$ |
| | 100% | 222.6 | 64×64 | 2.9 GB | $9.05 \times e^{998}$ |
| CIFAR100 | 0% (Original) | – | 32×32 | 737.6 MB | – |
| | 25% | 98.8 | 40×40 | 1.2 GB | $6.86 \times e^{452}$ |
| | 50% | 143.2 | 48×48 | 1.7 GB | $1.21 \times e^{686}$ |
| | 75% | 195.2 | 56×56 | 2.3 GB | $9.86 \times e^{858}$ |
| | 100% | 254.2 | 64×64 | 2.9 GB | $9.05 \times e^{998}$ |
| Imagenette | 0% (Original) | – | 224×224 | 7.5 GB | – |
| | 25% | 1110.6 | 280×280 | 11.8 GB | $9.58 \times e^{22245}$ |
| | 50% | 1594.6 | 336×336 | 16.9 GB | $4.54 \times e^{33679}$ |
| | 75% | 2146.8 | 392×392 | 23.0 GB | $1.62 \times e^{42154}$ |
| | 100% | 2866.8 | 448×448 | 30.1 GB | $3.39 \times e^{49013}$ |
| WikiText2 | 0% (Original) | – | – | 16.4 MB | – |
| | 25% | 34.21 | – | 20.5 MB | 53130 |
| | 50% | 41.4 | – | 24.6 MB | $3.01 \times e^{7}$ |
| | 75% | 47.98 | – | 28.7 MB | $3.24 \times e^{9}$ |
| | 100% | 54.82 | – | 32.8 MB | $1.37 \times e^{11}$ |
| AGNews | 0% (Original) | – | – | 138.2 MB | – |
| | 25% | 56.43 | – | 172.8 MB | $9.73 \times e^{37}$ |
| | 50% | 67.02 | – | 207.4 MB | $2.94 \times e^{58}$ |
| | 75% | 78.11 | – | 241.9 MB | $2.78 \times e^{73}$ |
| | 100% | 89.82 | – | 276.5 MB | $2.33 \times e^{86}$ |

**Results:** In Table 2, we present the augmentation results of the image datasets. The augmentation amount essentially corresponds to the percentage of augmentation (increase) in the dimensions of each image. Our results show that the dataset augmentation process lasts about up to a minute when it comes to datasets with 60K single-channel (grayscale) images, such as MNIST. It takes less than four minutes even when it comes to datasets with 60K three-channel (RGB) larger images, such as CIFAR10 and CIFAR100. When it comes to datasets, such as Imagenette, which contain higher resolution images, the dataset augmentation process takes longer. However, with higher-resolution images, lower augmentation amounts may be adequate as the search space grows exponentially even for smaller augmentation amounts (25% or even less). To this end, the dataset augmenter needs about 18 minutes for 25% of augmentation and about 9 minutes for 10% of augmentation (not shown in Table 2) to complete the dataset augmentation process for Imagenette.

We also present the text dataset augmentation results in Table 2. Text datasets are simpler to augment than image datasets as they consist of single-dimensional data and are easier to process because the tensors are integers. As such, the dataset augmenter can augment text datasets faster than image datasets (in less than 90 seconds). However, because of the simpler nature of text datasets, the search space is also smaller. As a result, higher augmentation amounts are needed for training datasets consisting of single-dimensional data.

## 5.3 NN Model Augmenter Evaluation

**Setup:** To evaluate the model augmenter for computer vision applications, we used four popular computer vision model families: ResNet [18], VGG [39], DenseNet [22], and MobileNet [20]. From each family, we selected a model that was augmented with amounts 25%, 50%, 75%, and 100%, and random sub-networks. Augmented models were trained with the corresponding augmented datasets used in Section 5.2. For NLP applications, we used two NLP models: a text classification model (consisting of an embedding layer and a fully connected layer) and a transformer network [44]. We augmented these models with amounts 25%, 50%, 75%, and 100%, and random sub-networks. The text classification model was trained with the augmented AGNews text dataset and the transformer model was trained with the augmented WikiText2 dataset used in Section 5.2. We quantified the number of parameters and training time before and after augmentation in addition to the training loss and accuracy. We also evaluated the fine-tuning ability of *Amalgam* by fine-tuning augmented pre-trained VGG16 models with augmented Imagenette datasets. For all experiments, we measured the training loss and accuracy for different augmentation amounts and compared them to the training loss and accuracy achieved by training the original (non-augmented) model with the original (non-augmented) dataset.

**ResNet results:** ResNet is a popular family of CNN-based computer vision models with residual blocks and skip connections. We used the ResNet-18 architecture for our evaluation. Figures 5a and 5b present the evaluation results (training loss and accuracy respectively) of training four separately augmented ResNet-18 models with four separately augmented MNIST datasets. The results suggest that even after augmentation, the model training is not affected, and the loss converges in the same manner as when the original model is trained with the original dataset. The same conclusions can be drawn when training the augmented ResNet-18 models with the augmented CIFAR10 (Figures 6a and 6b) and CIFAR100 datasets (Figures 7a and 7b). In Table 3, we present results for augmented ResNet-18 models in terms of model size (number of parameters) and training time. Our results indicate that as we increase the augmentation amount for a model and we train it with the corresponding augmented dataset, the model size and training time increase. To this end, users can select the desired model and dataset augmentation amounts based on their needs and constraints (*e.g.*, available time for training, and available cloud computing resources).

**VGG results:** VGG is another popular family of CNN-based models for computer vision. We used VGG16 for our evaluation. We augmented VGG16 with amounts 25%, 50%, 75%, and 100%, we trained those augmented models with datasets

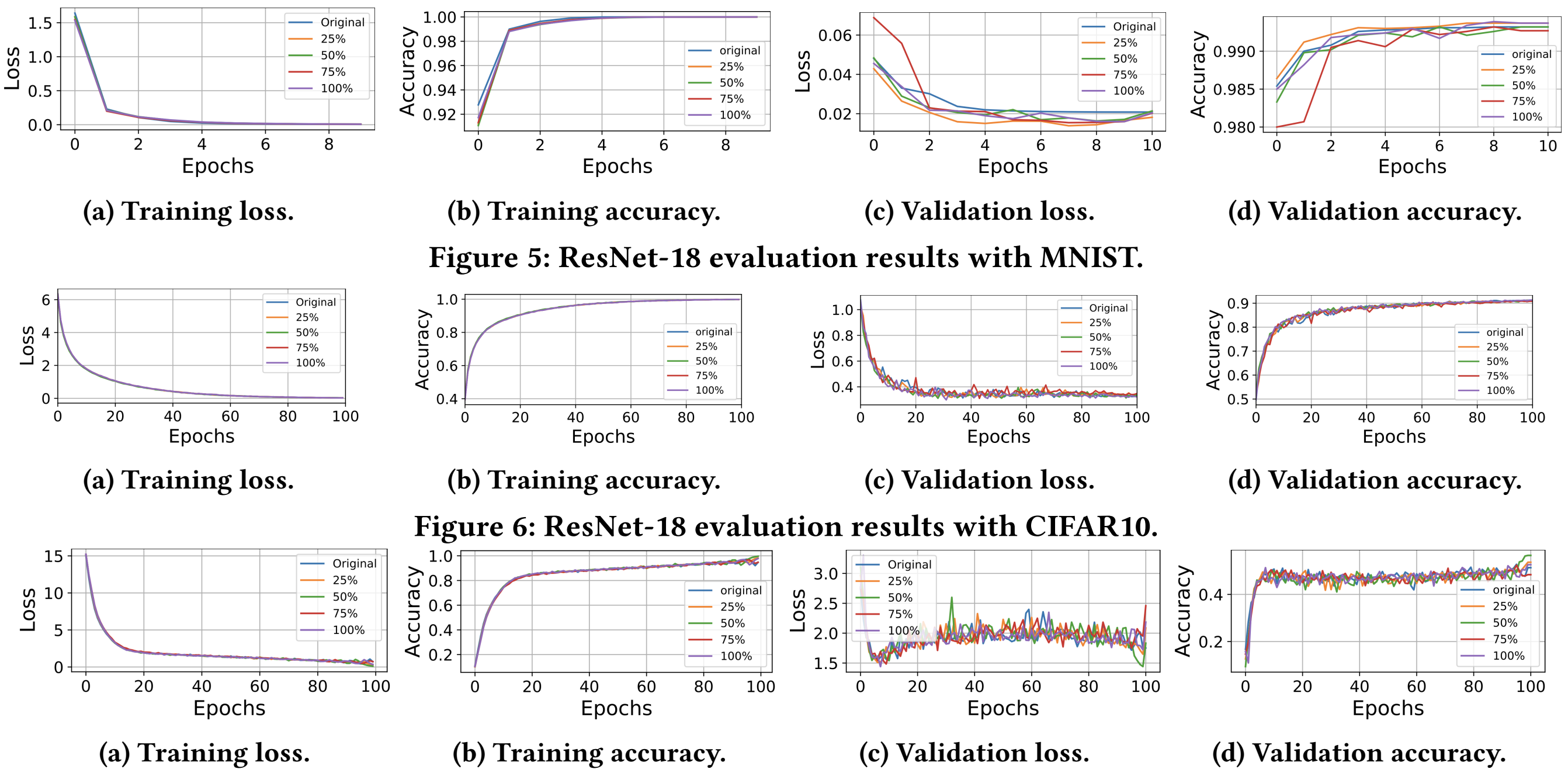

(a) Training loss. (b) Training accuracy. (c) Validation loss. (d) Validation accuracy.

**Figure 5: ResNet-18 evaluation results with MNIST.**

(a) Training loss. (b) Training accuracy. (c) Validation loss. (d) Validation accuracy.

**Figure 6: ResNet-18 evaluation results with CIFAR10.**

(a) Training loss. (b) Training accuracy. (c) Validation loss. (d) Validation accuracy.

**Figure 7: ResNet-18 evaluation results with CIFAR100.**

augmented by the same amount, and we compared their training loss and accuracy to the the training loss and accuracy of the original model when trained with the original dataset. In Figures 8a and 8b, we present the training loss and accuracy results respectively when the models are trained with MNIST. Our results indicate that the training of VGG16 is not affected by the model or dataset augmentation and the training loss converges in the same manner as when the original VGG16 model is trained with the original MNIST dataset. The same conclusions can be drawn when training the augmented VGG16 model with the augmented CIFAR10 (Figures 9a and 9b) and CIFAR100 datasets (Figures 10a and 10b). In Table 3, we present results for the number of parameters and training time of the augmented VGG16 models when trained with augmented datasets. Our results indicate that as we increase the augmentation amounts, the number of model parameters and the required training time increase.

Results of additional experiments with DenseNet121 and MobilenetV2 are presented in Appendix A.1.

**NLP results:** In Figure 11a, we present the training loss of the transformer model when trained with the WikiText2 dataset for different augmentation amounts. The results show that the training loss converges over time. Our transformer model predicts what the next word of a sentence should be (selects as the next word the word with the minimum loss). In this sense, the accuracy is not a meaningful metric for this model and that is the reason we present results only for training loss.

In Figures 12a and 12b, we present the training loss and accuracy results for the text classification model when trained with the AGNews dataset and different augmentation amounts. Our results indicate that the training process is not impacted by augmentation and the loss converges in the same manner as when the original model is trained with the original dataset.

In Table 4, we present the number of model parameters and the average training time for different amounts of model and dataset augmentation. Our results demonstrate the same trend as the results of Table 3 for the augmented computer vision models. The number of model parameters and the training time increase as we increase the augmentation amount.

**Transfer learning results:** To evaluate a fine-tuning scenario with *Amalgam*, we selected a VGG16 model with pre-trained weights from the ImageNet dataset [8]. We modified the pre-trained model by inserting Convolutional Block Attention Modules (CBAMs) [47]. Subsequently, we augmented the modified model with amounts 25%, 50%, 75%, and 100% and trained it with the corresponding augmented Imagenette datasets. Figure 13 presents the transfer learning evaluation results. The results demonstrate that the augmentation of the pre-trained model did not affect the fine-tuning of this model, and the training loss and accuracy results converge to the training results of the original pre-trained model.

## 5.4 NN Model Extractor Evaluation

**Setup:** The NN model extractor component extracts the original model from the augmented one once training of

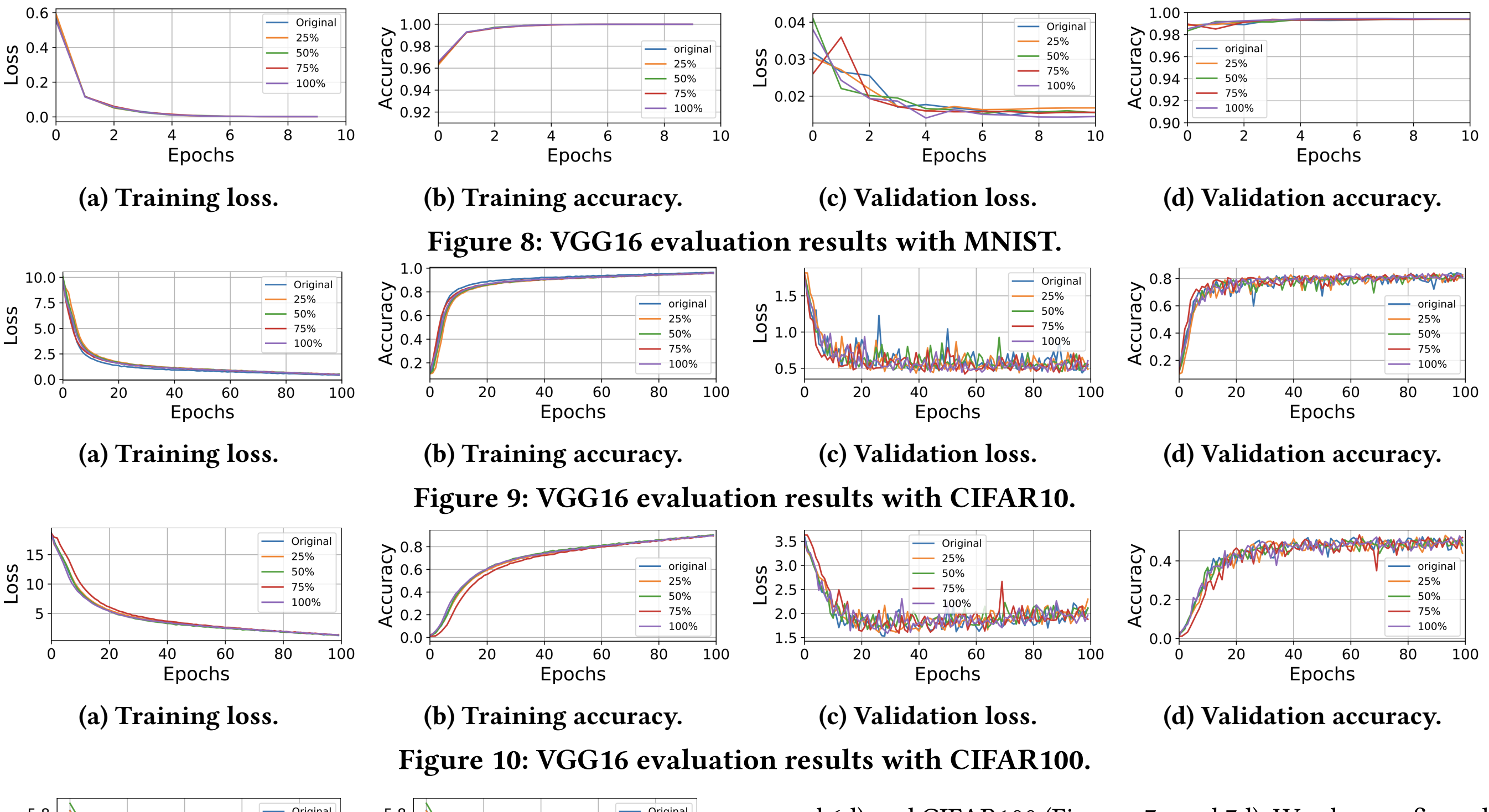

(a) Training loss. (b) Training accuracy. (c) Validation loss. (d) Validation accuracy.

**Figure 8: VGG16 evaluation results with MNIST.**

(a) Training loss. (b) Training accuracy. (c) Validation loss. (d) Validation accuracy.

**Figure 9: VGG16 evaluation results with CIFAR10.**

(a) Training loss. (b) Training accuracy. (c) Validation loss. (d) Validation accuracy.

**Figure 10: VGG16 evaluation results with CIFAR100.**

**(a) Model training loss with WikiText2.** **(b) Model validation loss with WikiText2.**

**Figure 11: NLP transformer model training with different augmentation amounts.**

the augmented model completes on the cloud. To evaluate the effectiveness of this component, we first validated the augmented model with the augmented testset and validated the de-obfuscated (extracted) model with the original testset. We further compared these validation results (validation loss and accuracy) with the validation results for the original (non-augmented) model trained with the original (non-augmented) training set and validated with the original (non augmented) testset. We evaluated the augmented models and datasets with augmentation amounts of 25%, 50%, 75%, and 100%.

**ResNet results:** The ResNet-18 model extraction results show no substantial difference in terms of validation loss and accuracy between validating the augmented model with the augmented testset and validating the de-obfuscated model with the original testset. Figures 5c and 5d present the validation results for MNIST indicating 0.07% of difference between the validation accuracy of the de-obfuscated model and the validation accuracy of the augmented models. The trend is the same for the validation results for CIFAR 10 (Figures 6c and 6d) and CIFAR100 (Figures 7c and 7d). We also confirmed that the validation loss and accuracy of the de-obfuscated model when validated with the original testset and the validation loss and accuracy of the original model when validated with the original testset show no difference.

**VGG results:** The VGG16 extraction results show no substantial difference in terms of validation loss and accuracy when we validated the augmented model with the augmented testset and when we validated the de-obfuscated model with the original testset. Figures 8c and 8d present the validation results for MNIST. These results indicate that there is 0.05% of difference between the validation accuracy of the de-obfuscated model and the validation accuracy of the augmented models. The trend is the same for the validation results for CIFAR 10 (Figures 9c and 9d) and CIFAR100 (Figures 10c and 10d). We further confirmed that the validation loss and accuracy of the de-obfuscated model when validated with the original testset and the validation loss and accuracy of the original model when validated with the original testset show no difference.

Results of additional experiments with DenseNet121 and MobilenetV2 are presented in Appendix A.2.

**NLP results:** In Figure 11b, we present the validation loss for the transformer model when validated with the WikiText2 testset. Our results show that the validation loss converges for all augmentation amounts. In Figures 12c and 12d, we present the validation loss and accuracy results respectively for the text classification model when validated with the AGNews testset. Our results show less than 2.0% of difference

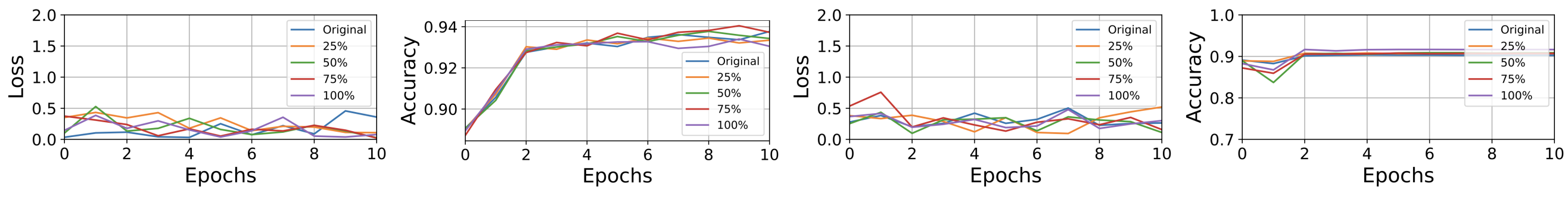

**(a) Text classification training loss with AGNews.** **(b) Text classification training accuracy with AGNews.** **(c) Text classification validation loss with AGNews.** **(d) Text classification validation accuracy with AGNews.**

**Figure 12: NLP text classification model training with different augmentation amounts.**

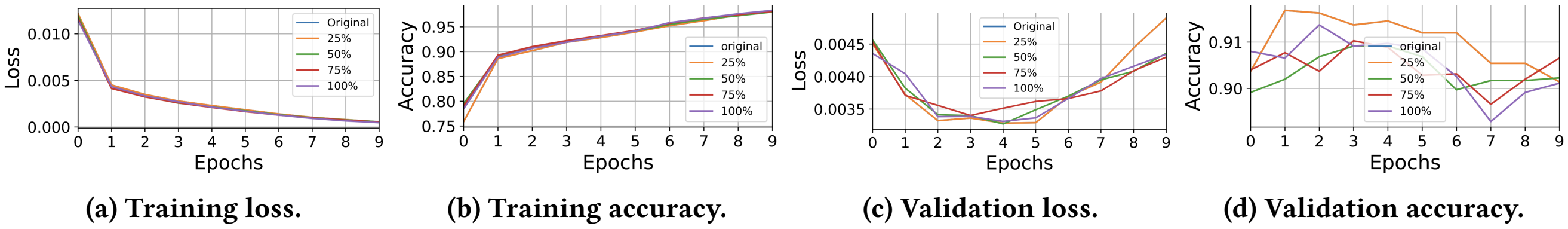

**(a) Training loss.** **(b) Training accuracy.** **(c) Validation loss.** **(d) Validation accuracy.**

**Figure 13: Transfer learning evaluation results with VGG16 and Imagenette.**

**Table 3: Computer vision model training with different augmentation amounts.**

| Dataset | Augmentation amount | Number of parameters after augmentation | | | |
|---|---|---|---|---|---|
| | | ResNet18 | VGG16 | DenseNet121 | MobileNetV2 |
| MNIST | 0% (Original) | $11.17 \times 10^6$ | $14.72 \times 10^6$ | $10.00 \times 10^5$ | $22.96 \times 10^5$ |
| | 25% | $13.96 \times 10^6$ | $18.40 \times 10^6$ | $12.50 \times 10^5$ | $28.70 \times 10^5$ |
| | 50% | $17.09 \times 10^6$ | $22.09 \times 10^6$ | $15.01 \times 10^5$ | $34.44 \times 10^5$ |
| | 75% | $19.60 \times 10^6$ | $25.77 \times 10^6$ | $17.50 \times 10^5$ | $40.18 \times 10^5$ |
| | 100% | $22.34 \times 10^6$ | $29.45 \times 10^6$ | $20.00 \times 10^5$ | $45.92 \times 10^5$ |
| CIFAR10 | 0% (Original) | $11.17 \times 10^6$ | $14.72 \times 10^6$ | $10.00 \times 10^5$ | $22.96 \times 10^5$ |
| | 25% | $13.97 \times 10^6$ | $18.40 \times 10^6$ | $12.50 \times 10^5$ | $28.70 \times 10^5$ |
| | 50% | $17.09 \times 10^6$ | $22.09 \times 10^6$ | $15.00 \times 10^5$ | $34.44 \times 10^5$ |
| | 75% | $19.60 \times 10^6$ | $25.77 \times 10^6$ | $17.50 \times 10^5$ | $40.18 \times 10^5$ |
| | 100% | $22.34 \times 10^6$ | $29.45 \times 10^6$ | $20.00 \times 10^5$ | $45.93 \times 10^5$ |
| CIFAR100 | 0% (Original) | $11.22 \times 10^6$ | $14.77 \times 10^6$ | $10.35 \times 10^5$ | $24.12 \times 10^5$ |
| | 25% | $14.04 \times 10^6$ | $18.40 \times 10^6$ | $12.50 \times 10^5$ | $28.70 \times 10^5$ |
| | 50% | $17.17 \times 10^6$ | $22.09 \times 10^6$ | $15.00 \times 10^5$ | $34.44 \times 10^5$ |
| | 75% | $19.68 \times 10^6$ | $25.77 \times 10^6$ | $17.50 \times 10^5$ | $40.18 \times 10^5$ |
| | 100% | $22.44 \times 10^6$ | $29.45 \times 10^6$ | $20.70 \times 10^5$ | $48.24 \times 10^5$ |
| | | Custom VGG16 with CBAM modules | | | |
| Imagenette | 0% (Original) | $13.84 \times 10^7$ | | | |
| | 25% | $17.29 \times 10^7$ | | | |
| | 50% | $20.75 \times 10^7$ | | | |
| | 75% | $24.21 \times 10^7$ | | | |
| | 100% | $27.67 \times 10^7$ | | | |

| Dataset | Augmentation amount | Average training time (s) | | | |
|---|---|---|---|---|---|
| | | ResNet | VGG16 | DenseNet121 | MobileNetV2 |
| MNIST | 0% (Original) | 164.6 | 207.1 | 547.9 | 324.7 |
| | 25% | 207.0 | 519.4 | 977.8 | 510.7 |
| | 50% | 217.1 | 834.7 | 990.3 | 528.9 |
| | 75% | 226.4 | 1082.8 | 1000.1 | 540.7 |
| | 100% | 233.2 | 1371.5 | 1003.9 | 540.9 |
| CIFAR10 | 0% (Original) | 1358.8 | 1707.6 | 4804.4 | 2862.2 |
| | 25% | 2755.7 | 4813.8 | 7405.3 | 4431.1 |
| | 50% | 3059.4 | 7529.8 | 7640.1 | 4624.1 |
| | 75% | 3287.0 | 9737.9 | 7769.9 | 4913.4 |
| | 100% | 3473.9 | 12081.0 | 7912.2 | 4939.1 |
| CIFAR100 | 0% (Original) | 1764.1 | 1713.2 | 4721.1 | 2559.9 |
| | 25% | 2528.1 | 4841.8 | 7409 | 4605 |
| | 50% | 2849.6 | 7560.6 | 7642.3 | 4801.8 |
| | 75% | 3079.4 | 9744.8 | 7776.7 | 4888.9 |
| | 100% | 3279.5 | 12145.7 | 7915.9 | 4980.6 |
| | | Custom VGG16 with CBAM modules | | | |
| Imagenette | 0% (Original) | 548.2 | | | |
| | 25% | 754.6 | | | |
| | 50% | 859.34 | | | |
| | 75% | 964.2 | | | |
| | 100% | 1084.3 | | | |

**Table 4: NLP model training with different augmentations.**

| Model/Dataset | Augmentation amount | Parameters | Training time (s) |
|---|---|---|---|
| Transformer model/WikiText2 | 0% (Original) | $12.03 \times 10^6$ | 375.7 |
| | 25% | $15.03 \times 10^6$ | 405.4 |
| | 50% | $18.04 \times 10^6$ | 436.9 |
| | 75% | $21.04 \times 10^6$ | 498.2 |
| | 100% | $24.05 \times 10^6$ | 520.3 |
| Text classification model/AGNews | 0% (Original) | $6.13 \times 10^6$ | 69.7 |
| | 25% | $7.67 \times 10^6$ | 72.4 |
| | 50% | $9.19 \times 10^6$ | 78.9 |
| | 75% | $10.73 \times 10^6$ | 83.7 |
| | 100% | $12.26 \times 10^6$ | 90.3 |

between the validation accuracy of the de-obfuscated model and the validation accuracy of the augmented models. We confirmed that the validation loss and accuracy of the de-obfuscated model when validated with the original testset and the validation loss and accuracy of the original model when validated with the original testset show no substantial difference.

**Transfer learning results:** The pre-trained VGG16 model shows the same trend in terms of validation loss and accuracy when validated with the original Imagenette testset (Figure 13). Among different augmentation amounts, there is only a 0.4% difference in terms of validation accuracy.

**Miscellaneous results:** Our evaluation indicates that *Amalgam* does not increase inference time for de-obfuscated models. Specifically, the inference time for de-obfuscated models is the same as the inference time of the original models. This is because a de-obfuscated model has the same number of parameters as the original one. The extraction of a model happens in constant time (typically it takes a few milliseconds) and does not depend on the augmentation amount.

## 5.5 Performance Comparison With Other Frameworks

To evaluate the performance of *Amalgam* in comparison with other privacy-preserving NN model training approaches, we have compared *Amalgam* with PyCrCNN [10], CrypTen [23], and DISCO [40]. PyCrCNN is based on FHE implemented in PyTorch. It is essentially a distributed architecture for secure deep-learning-as-a-service tailored towards CNNs. CrypTen is an MPC-based privacy-preserving deep-learning training approach for modular NNs. DISCO is a dynamic obfuscation framework for DNNs. We trained the LeNet [27] model using the MNIST dataset for 10 epochs with a learning rate of 0.001 and a batch size of 128. For *Amalgam*, we have used 100% augmentation for the model and the dataset. For PyCrCNN, we have used $2^{11}$ as the polynomial modulus degree and 37780 as the plaintext modulus. Finally, we have used three parties for CrypTen. We have compared the performance of these frameworks with vanilla PyTorch (no obfuscation) as our baseline and with training on a CPU (Intel Core i9 10900K). Model training based on CPU is used to represent a TEE-based solution limited to the use of CPU for training, such as TensorScone [26], under the assumption that the TEE does not incur additional overhead (it is essentially the best-case scenario for such a TEE-based solution).

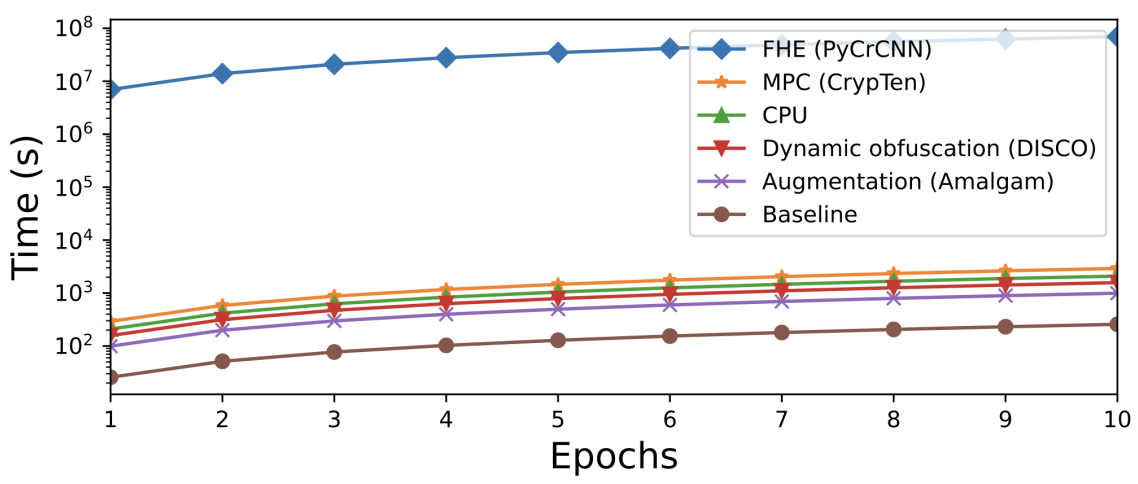

**Figure 14: LeNet training time comparison of *Amalgam* with other frameworks for privacy-preserving machine learning.**

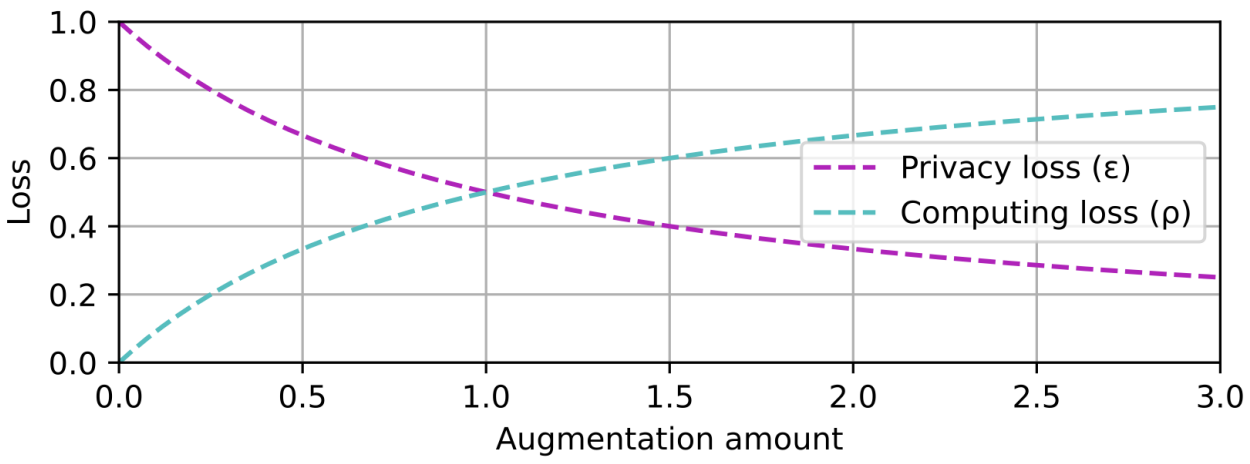

**Figure 15: Privacy and computing performance loss for different augmentation amounts.**

Figure 14 presents the training times for all frameworks. With no privacy preservation, one epoch took 25 seconds on average. With *Amalgam*, it took 1 minute and 39 seconds. It took 2 min 38 sec with DISCO (36% slower). With CrypTen, it took 4 minutes and 52 seconds, which is 11× slower than the baseline and about 3× slower than *Amalgam*. With CPU only, training was 8× and 2× slower than the baseline and *Amalgam* respectively. Finally, PyCrCNN took over 3 days to complete, making it 13440× slower than the baseline and about 3393× slower than *Amalgam*. Moreover, FHE has a limitation that prevents the execution of non-linear operations. As such, the last layer of the LeNet model was replaced with a simple squared function in the case of PyCrCNN. To this end, PyCrCNN reached a validation accuracy of 95%, while the baseline, *Amalgam*, CPU, and CrypTen reached a validation accuracy of 98%.

# 6 SECURITY AND PRIVACY ANALYSIS

## 6.1 Privacy Loss

Let us consider the dataset augmenter component where $X \in \{x_1, x_2, ..., x_n\}$ is an original sample with $n$ features and $X' \in \{x_1, x_2, ..., x_n, x'_{n+1}\}$ is an augmented sample with $n+1$ features (an extra arbitrary feature $x'_{n+1}$ is added). Let $\mathcal{A}$ be an adversary that performs a randomized query $q$ to retrieve a feature $s$. If we compare the probabilities of $\mathcal{A}$ finding an original feature between $X$ and $X'$ through $q$, we have:

$$\Pr[s \in X : s \leftarrow \mathcal{A}^{q(X)}] > \Pr[s \in X : s \leftarrow \mathcal{A}^{q(X')}]. \quad (3)$$

However, if another adversary $\mathcal{A}'$ has access to $m$ information $\{a_1, a_2, ..., a_m\} \subset X'$, then $\mathcal{A}'$ can perform an advanced query $q_a$, and the probability becomes:

$$\Pr[s \in X : s \leftarrow \mathcal{A}^{q(X')}] < \Pr[s \in X : s \leftarrow \mathcal{A}'^{q_a(X')}]. \quad (4)$$

In other words, the more augmentation we add, the less likely it is for $\mathcal{A}$ and $\mathcal{A}'$ to find an original data sample. As such, we can quantify the privacy loss $\epsilon$ for an augmentation amount $\alpha$ from Equation 4 using:

$$\epsilon = 1/(1+\alpha). \quad (5)$$

The same analogy applies for model augmentation as well since the addition of more arbitrary parameters will better hide the original parameters.

## 6.2 Computing Performance Loss

Although augmentation has no direct impact on accuracy, it incurs computing overhead. Subsequently, the computing performance loss $\rho$ for an augmentation amount $\alpha$ can be quantified as follows:

$$\rho = 1 - [1/(1+\alpha)]\,. \quad (6)$$

The above privacy loss and computing performance loss analysis also applies to model augmentation, where the number of augmented parameters impacts model privacy and computing performance. Figure 15 presents the privacy and computing performance loss for various augmentation amounts. Using Equations 5 and 6, users can determine how much privacy they can afford at the cost of computing overhead.

## 6.3 Adversarial Attack Analysis

We performed different adversarial attacks to analyze the resilience of *Amalgam*. This analysis helped us identify not only the privacy guarantees provided by *Amalgam* but also the impact of different augmentation levels on different attacks. We discuss various server-side attacks that can be performed against *Amalgam* below.

| Method | Plain cifar10 + LeNet | | Augmented cifar10 + Plain LeNet | |
|---|---|---|---|---|
| | Iteration=0 | Iteration=84 | Iteration=0 | Iteration=84 |
| DLG | | | | |
| iDLG | | | | |

**Figure 16: DLG attack on an augmented CIFAR10 image and LeNet model.**

**Brute-force attack:** When augmented with *Amalgam*, the level of augmentation increases the search space making it more and more difficult for the server to perform a successful brute-force attack. Table 2 presents the search space for different amounts of augmentation. The user can select augmentation amounts accordingly, so the number of brute force attempts an adversary needs to conduct remains high enough, thus making such an attack infeasible.

**Deep leakage from gradient attack:** Recent studies have shown that sharing gradients can leak enough information to reconstruct the training dataset [58]. Federated learning is susceptible to such attacks. Since the server is training the model, it has access to its gradients. *Amalgam* is not susceptible to such attacks because of two reasons: (i) the augmented model contains additional synthetic parameters (the adversary has no knowledge about the original parameters) that equally participate in gradient descent, and (ii) the entire augmented input goes into the augmented model making the reconstruction of the original input data infeasible for an adversary. To test this hypothesis and demonstrate the resilience of *Amalgam* against such an attack, we performed both a Deep Leakage from Gradient (DLG) [58] and an improved Deep Leakage from Gradient (iDLG) [55] attack on a 50% augmented LeNet model trained with a 50% augmented MNIST dataset. Figure 16 presents the results after 84 iterations showing the failure of reconstruction.

**Model Inversion attacks:** We performed an attack using a model explanation technique to investigate whether it is feasible to identify the original sub-network from an augmented model. We used SHAP [29] on LeNet trained with MNIST with 100% augmentation and three sub-networks. Figure 17 presents the SHAP values before and after augmentation. The results show that augmentation produced highly distorted SHAP values, which cannot be used to explain the model. Zhou et. al. showed that white-box gradient attacks like FGSM [17], BIM [41], and PGD [30] are ineffective when a model is obfuscated with parameter augmentation [57].

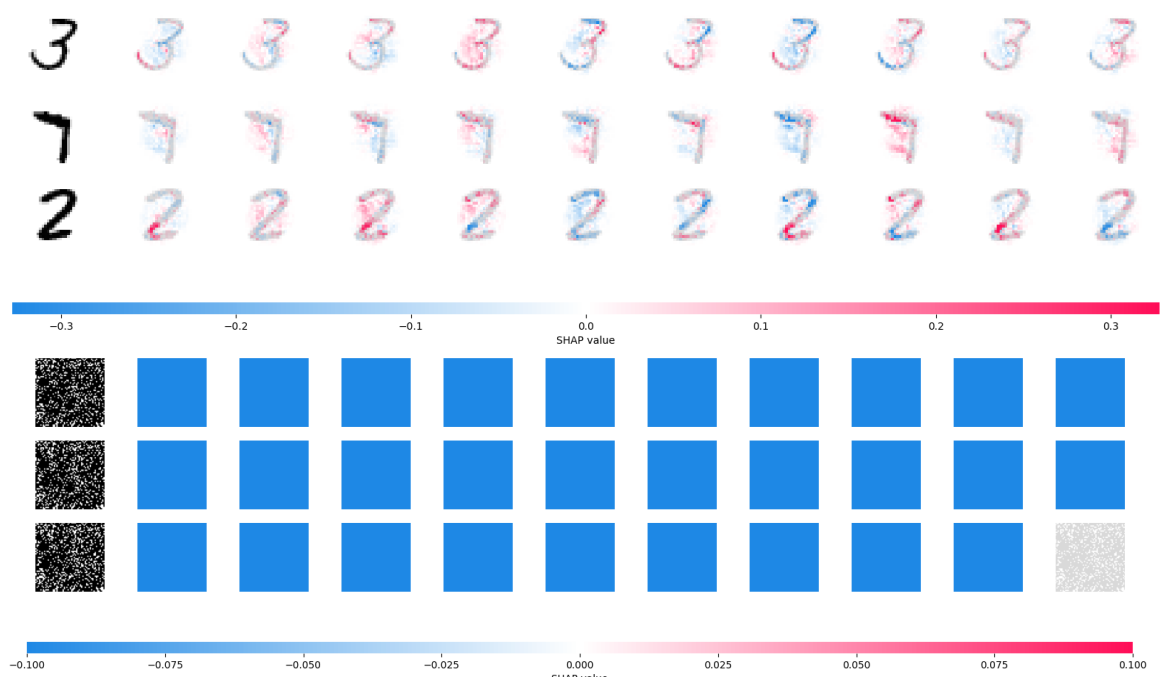

**Figure 17: Comparison of SHAP values before (top) and after (bottom) augmentation of LeNet.**

| Augmentation | Augmented image | Denoise with Restormer | Denoise with KBNet | Ground truth |
|---|---|---|---|---|
| Simple gaussian noise σ = 50 | | | | |
| Amalgam augmentation Type = guassian Amount = 20% σ = 50 | | | | |

**Figure 18: Denoising attack on an augmented image.**

**Deep denoising attack:** *Amalgam* hides the original input data in a dataset by augmenting them with synthetic information (noise). However, this is not just a simple noise (e.g., Gaussian noise). For example, when simple noise is added to an image, it randomly changes the value of some pixels while retaining the original form of the image. *Amalgam* augments synthetic pixel information in-between original pixels, which not only changes the dimensions but also distorts the image. As such, the original image data cannot be retrieved by denoising the augmented image. To test this hypothesis, we performed two state-of-the-art deep denoising attacks against *Amalgam* (Restormer [52] and KBNet [54]). Figure 18 presents the results showing that both approaches fail to denoise the augmented image with only 20% augmentation.

## 7 CONCLUSION AND FUTURE WORK

In this paper, we proposed *Amalgam*, a framework for obfuscated NN model training on the cloud. *Amalgam* achieves that by augmenting the model architecture and dataset to be used for training with synthetic information in ways that do not impact the training correctness of the original network. This comes at the cost of additional training time and computing overhead. In the future, we plan to extend our evaluation to include additional types of NN model architectures and datasets. Finally, we plan to enhance the design components of *Amalgam* to reduce the required augmentation time, training time, and computing overhead.

# A APPENDIX

## A.1 Additional NN Model Augmenter evaluation

**DenseNet results:** The DenseNet extraction results show no substantial difference in terms of validation loss and accuracy for the augmented and the de-obfuscated models. For the MNIST dataset, Figures 19c and 19d present the validation loss and accuracy results respectively, showing 1.05% of difference between the validation accuracy of the de-obfuscated model and the validation accuracy of the augmented models. The trend is the same for the validation results for CIFAR10 (Figures 20c and 20d) and CIFAR100 (Figures 21c and 21d). We further confirmed that the validation loss and accuracy of the de-obfuscated model when validated with the original testset and the validation loss and accuracy of the original model when validated with the original testset show no difference.

**MobileNet results:** The MobileNetV2 extraction results also show no substantial difference in terms of validation loss accuracy. Figures 22c and 22d present the validation results for MNIST, which indicate 1.05% of difference between the validation accuracy of the de-obfuscated model and the validation accuracy of the augmented models. The trend is the same for the validation results for CIFAR10 (Figures 23c and 23d) and CIFAR100 (Figures 24c and 24d). We confirmed that the validation loss and accuracy of the de-obfuscated model when validated with the original testset and the validation loss and accuracy of the original model when validated with the original testset show no substantial difference.

## A.2 Additional NN Model Extractor evaluation

**DenseNet results:** The DenseNet extraction results show no substantial difference in terms of validation loss and accuracy for the augmented and the de-obfuscated models. For the MNIST dataset, Figures 19c and 19d present the validation loss and accuracy results respectively, showing 1.05% of difference between the validation accuracy of the de-obfuscated model and the validation accuracy of the augmented models. The trend is the same for the validation results for CIFAR10 (Figures 20c and 20d) and CIFAR100 (Figures 21c and 21d). We further confirmed that the validation loss and accuracy of the de-obfuscated model when validated with the original testset and the validation loss and accuracy of the original model when validated with the original testset show no difference.

**MobileNet results:** The MobileNetV2 extraction results also show no substantial difference in terms of validation loss accuracy. Figures 22c and 22d present the validation results for MNIST, which indicate 1.05% of difference between the validation accuracy of the de-obfuscated model and the validation accuracy of the augmented models. The trend is the same for the validation results for CIFAR10 (Figures 23c and 23d) and CIFAR100 (Figures 24c and 24d). We confirmed that the validation loss and accuracy of the de-obfuscated model when validated with the original testset and the validation loss and accuracy of the original model when validated with the original testset show no substantial difference.

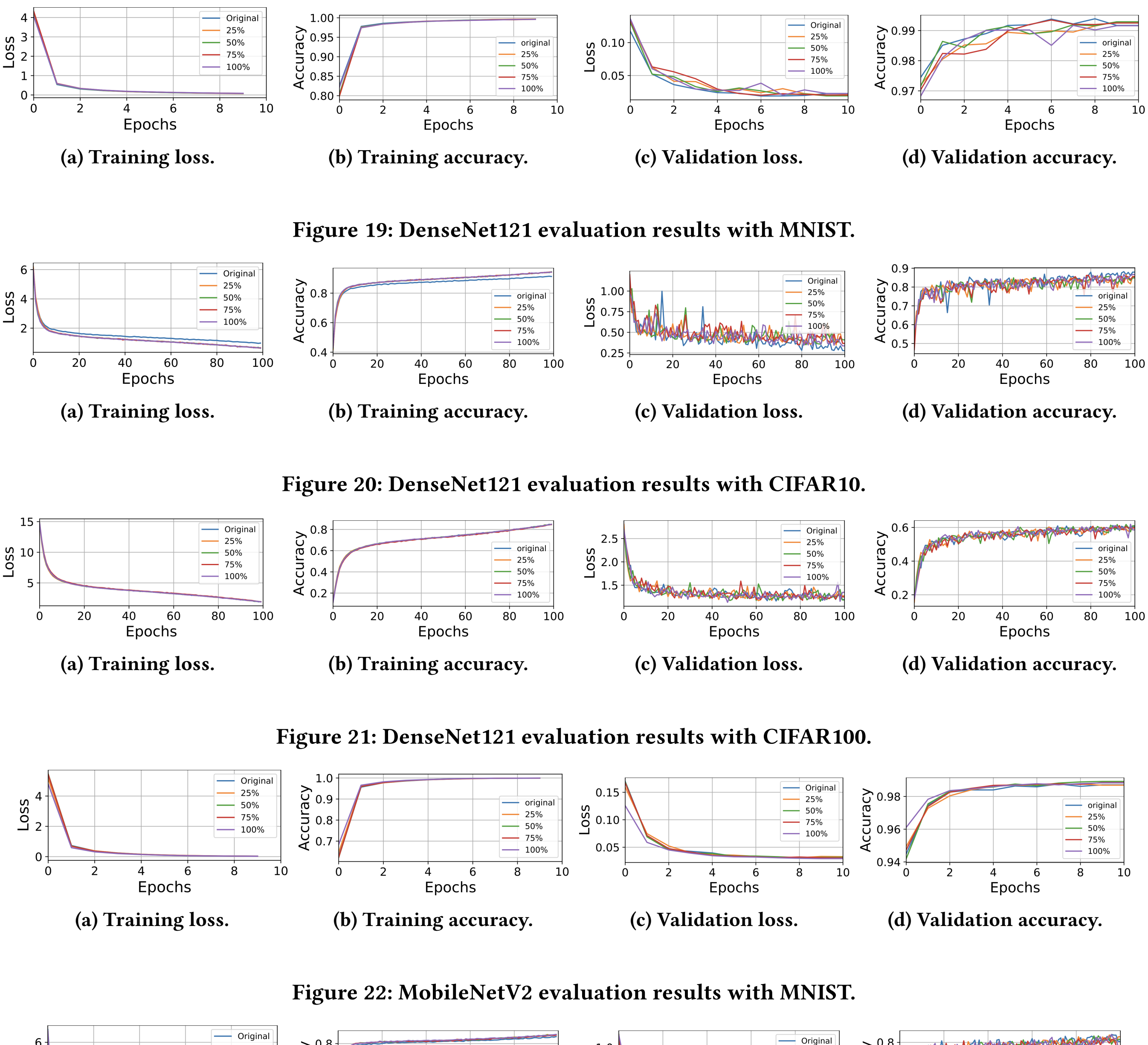

(a) Training loss. (b) Training accuracy. (c) Validation loss. (d) Validation accuracy.

**Figure 19: DenseNet121 evaluation results with MNIST.**

(a) Training loss. (b) Training accuracy. (c) Validation loss. (d) Validation accuracy.

**Figure 20: DenseNet121 evaluation results with CIFAR10.**

(a) Training loss. (b) Training accuracy. (c) Validation loss. (d) Validation accuracy.

**Figure 21: DenseNet121 evaluation results with CIFAR100.**

(a) Training loss. (b) Training accuracy. (c) Validation loss. (d) Validation accuracy.

**Figure 22: MobileNetV2 evaluation results with MNIST.**

(a) Training loss. (b) Training accuracy. (c) Validation loss. (d) Validation accuracy.

**Figure 23: MobileNetV2 evaluation results with CIFAR10.**

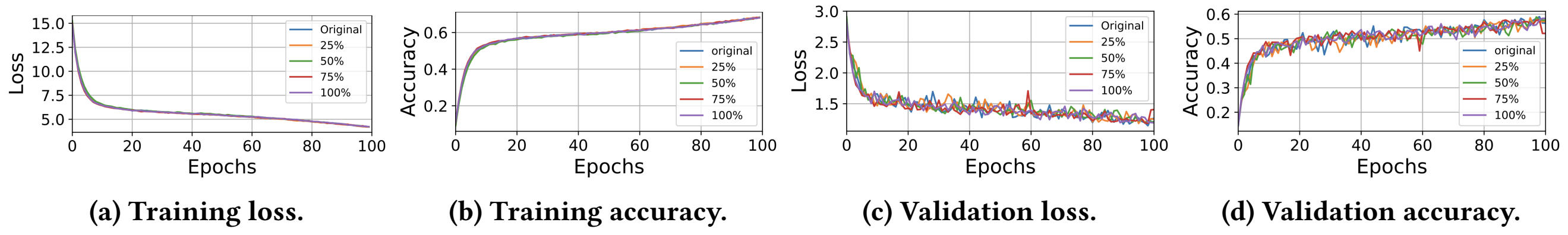

**(a) Training loss.** **(b) Training accuracy.** **(c) Validation loss.** **(d) Validation accuracy.**

**Figure 24: MobileNetV2 evaluation results with CIFAR100.**